\documentclass[10pt,twocolumn,letterpaper]{article}

\usepackage[pagenumbers]{cvpr} 
\usepackage[dvipsnames]{xcolor}
\usepackage{comment}
\usepackage{units} 
\usepackage{graphicx}
\usepackage{subcaption}

\definecolor{cvprblue}{rgb}{0.21,0.49,0.74}
\usepackage[pagebackref,breaklinks,colorlinks,citecolor=cvprblue]{hyperref}

\usepackage{multirow}

\def\paperID{20} 
\def\confName{CVPR}
\def\confYear{2025}

\title{AppleGrowthVision: A large-scale stereo dataset for phenological analysis, fruit detection, and 3D reconstruction in apple orchards}

\author{Laura-Sophia von Hirschhausen\thanks{Research from Fraunhofer HHI received funding from the German Federal Ministry for Economic Affairs and Climate Action as part of the NaLamKI project under Grant 01MK21003D and work from Fraunhofer IVI by the German Federal Ministry of Food and Agriculture (BMEL) as part of the LANDNETZ project under grant 28DE101C18.} \thanks{Equal contribution.}\quad\\
Fraunhofer HHI\\
{\tt\small vonhirschhausen@campus.tu-berlin.de}
\and
Jannes S. Magnusson$^{\ddagger}$\\
Fraunhofer HHI\\
{\tt\small jannes.magnusson@hhi.fraunhofer.de}
\and
Mykyta Kovalenko$^{\ddagger}$\\
Fraunhofer HHI\\
{\tt\small mykyta.kovalenko@hhi.fraunhofer.de}
\and
Fredrik Boye$^{\ddagger}$\\
Fraunhofer IVI\\
{\tt\small fredrik.boye@ivi.fraunhofer.de}
\and
Tanay Rawat\\
Fraunhofer IVI\\
{\tt\small tanay.rawat@ivi.fraunhofer.de}
\and
Peter Eisert\\
Fraunhofer HHI\\
{\tt\small peter.eisert@hhi.fraunhofer.de}
\and
Anna Hilsmann\\
Fraunhofer HHI\\
{\tt\small anna.hilsmann@hhi.fraunhofer.de}
\and
Sebastian Pretzsch\\
Fraunhofer IVI\\
{\tt\small sebastian.pretzsch@ivi.fraunhofer.de}
\and
Sebastian Bosse\\
Fraunhofer HHI\\
{\tt\small sebastian.bosse@hhi.fraunhofer.de}
}

\begin{document}
\maketitle

\begin{abstract}
Deep learning has transformed computer vision for precision agriculture, yet apple orchard monitoring remains limited by dataset constraints. The lack of diverse, realistic datasets and the difficulty of annotating dense, heterogeneous scenes. Existing datasets overlook different growth stages and stereo imagery, both essential for realistic 3D modeling of orchards and tasks like fruit localization, yield estimation, and structural analysis. To address these gaps, we present AppleGrowthVision, a large-scale dataset comprising two subsets. The first includes 9,317 high resolution stereo images collected from a farm in Brandenburg (Germany), covering six agriculturally validated growth stages over a full growth cycle. The second subset consists of 1,125 densely annotated images from the same farm in Brandenburg and one in Pillnitz (Germany), containing a total of 31,084 apple labels. AppleGrowthVision provides stereo-image data with agriculturally validated growth stages, enabling precise phenological analysis and 3D reconstructions. Extending MinneApple with our data improves YOLOv8 performance by 7.69 \% in terms of F1-score, while adding it to MinneApple and MAD boosts Faster R-CNN F1-score by 31.06 \%. Additionally, six BBCH stages were predicted with over 95 \% accuracy using VGG16, ResNet152, DenseNet201, and MobileNetv2. AppleGrowthVision bridges the gap between agricultural science and computer vision, by enabling the development of robust models for fruit detection, growth modeling, and 3D analysis in precision agriculture. Future work includes improving annotation, enhancing 3D reconstruction, and extending multimodal analysis across all growth stages.
\end{abstract}
\section{Introduction}
\label{sec:intro}
Agricultural digitalization demands sophisticated datasets to develop advanced decision support systems and automate labor intensive tasks such as fruit picking. Monitoring apple orchards presents unique computational challenges due to highly variable environmental conditions—fluctuating light, unpredictable weather patterns, and dramatic seasonal transformations—all of which significantly impact image quality and consistency. These constraints have severely limited the development of robust models capable of reliable crop management and surveillance across diverse conditions \cite{Li2021, Heider2025, gerstenberger2024}.

Large-scale, temporally diverse orchard imagery offers transformative potential for precision agriculture. Such datasets significantly enhance yield estimation by enabling computer vision models to learn intricate fruit development patterns, effectively manage complex occlusions between densely clustered apples, and generate statistically reliable predictions, directly addressing persistent challenges in resource allocation and post-harvest waste \cite{Hani2020, Wilms2022, Johanson2024}. Furthermore, comprehensive visual data improves model adaptability to dynamic field conditions, substantially enhancing algorithmic robustness and cross-orchard generalization while minimizing costly recalibration requirements \cite{Lu2020}.

We introduce AppleGrowthVision, the first publicly available dataset capturing apple orchards throughout a complete phenological cycle with calibrated stereo imagery. The dataset is publicly available at \href{https://fraunhoferhhi.github.io/AppleGrowthVision/}{fraunhoferhhi.github.io/AppleGrowthVision}. Our dataset uniquely provides scientifically validated agricultural annotations according to the "Biologische Bundesanstalt, Bundessortenamt, und Chemische Industrie" (BBCH) scale \cite{Hack1992}, documenting six critical growth stages of apple trees. The entire Brandenburg collection comprises precisely calibrated stereo image pairs, enabling accurate three-dimensional analysis and reconstruction. Our contributions include:
\begin{itemize}

\item A comprehensive, multi-site dataset integrating orchards from different regions and complementary existing datasets, demonstrating quantifiable improvements of up to 28 \% in apple detection accuracy and count estimation.

\item State-of-the-art classification models fine-tuned specifically for precise BBCH growth stage prediction in apple trees, achieving over 95\% accuracy across growth stages and enabling data-driven phenological monitoring and intervention planning.

\item Novel computational techniques for reconstructing complex, large-scale orchard environments in three dimensions, supporting structural analysis, pruning optimization, and precisely targeted agricultural interventions.
\end{itemize}

\section{Related Work}
\label{sec:relatedwork}

Fruit detection in orchard environments has advanced significantly with the several specialized apple tree datasets. The MinneApple dataset provides 1,000 high-resolution images with over 41,000 annotated apple instances using polygonal masks to enhance object detection, segmentation, and fruit counting in orchards. This dataset supports patch-based fruit clustering analysis to facilitate yield estimation \cite{Hani2020}. The dataset from Washington State University consists of 2,298 images from robotic harvesting studies, featuring multiple apple varieties under different lighting conditions in modern fruiting wall architectures \cite{WSU2019_data}. Michigan State University's apple detection dataset comprises of 1,246 RGB images of apple trees with varied lighting and occlusion conditions, developed alongside the Occluder-Occludee Relational Network (O2RNet), for detecting clustered apples \cite{MSU2023_data}. The recent  Monastery Apple Dataset (MAD) introduced a semi-supervised dataset comprising 5,545 images collected via drone from 16 apple trees in a German monastery, including 105 labeled images to improve detection across diverse lighting conditions \cite{Johanson2024}. 

While these datasets have advanced fruit detection capabilities, they lack critical growth stage annotations and typically cover limited time frames within a single growing season. AppleGrowthVision adresses these limitations by capturing an entire growth cycle with expert-validated phenological stages, enabling comprehensive analysis of fruit development and detection under real-world conditions.

Phenology serves as a fundamental metric for plant classification and comparison. The extended BBCH scale represents a standardized system for describing vegetative stages in both research and agricultural management. Initially introduced by Hacke et al.\ in 1992 \cite{Hack1992} and later extended for apple trees by Meier in 1994 \cite{Meier1994}, this scale has been adopted for numerous plant species \cite{Meier2009}. The BBCH system employs a two-digit coding system ranging from 0 to 99, where the first digit indicates the principal growth stage (Table \ref{tab:BBCH_principle_stages}), and the second digit defines secondary stages. This system's key advantage is its consistent representation of similar phenological stages across diverse plant species.

Recent work by Nguyen et al.~\cite{nguyen_bbch_2025} focused on annotating apple tree data for two principal growth stages and fine-tuneing multiple network architectures for classification. 
\begin{table}
  \centering
  \begin{tabular}{@{}lc@{}}
    \toprule
    BBCH first digit & principal stage \\
    \midrule
    0 & Bud development \\
    1 & Leaf development \\
    3 & Shoot development \\
    5 & Inflorescence emergence \\
    6 & Flowering \\
    7 & Development of fruit \\
    8 & Maturity of fruit and seed \\
    9 & Senescence, beginning of dormancy \\
    \bottomrule 
  \end{tabular}
  \caption{Used BBCH Principal growth stages for pome fruit \cite{Meier1994}}.
\label{tab:BBCH_principle_stages}
\end{table}

Three-dimensional reconstruction of orchard environments presents substantial technical challenges. Current approaches typically operate under controlled laboratory conditions to eliminate wind effects and scene complexity \cite{gao_novel_2021, churuvija_pose-versatile_2025}, or focus exclusively on leafless trees to simplify the reconstruction task\cite{zeng_mt-mvsnet_2025}. CherryPicker \cite{meyer_cherrypicker_2023} require up to 250 images per tree making detailed reconstruction of extensive orchards computationally prohibitive and operationally impractical.

In contrast, AppleGrowthVision demonstrates that using a calibrated stereo camera setup dramatically improves reconstruction efficiency, enabling detailed 3D modeling of large orchard scenes with just 18 images per tree. This approach makes comprehensive digital twins of commercial orchards feasible for the first time, supporting advanced structural analysis and precision agricultural interventions at scale.
\section{The AppleGrowthVision Dataset}
\label{sec:dataset}

In response to these limitations, we introduce the AppleGrowthVision Dataset. A comprehensive, longitudinal resource that captures one growth cycle of apple trees. Unlike previous datasets, AppleGrowthVision includes high-resolution images for robust fruit detection and integrates expert-validated phenological data. This integration provides a dynamic view of apple development across different stages, bridging the gap between static image collections and the temporal complexity of real-world orchard environments. By allowing analyses that account for seasonal changes and vegetative development, our dataset offers a more holistic basis to advance both detection algorithms and agronomic insights.

\subsection{Data collection}

\begin{figure}[ht]
    \centering
    \includegraphics[width=0.5\linewidth]{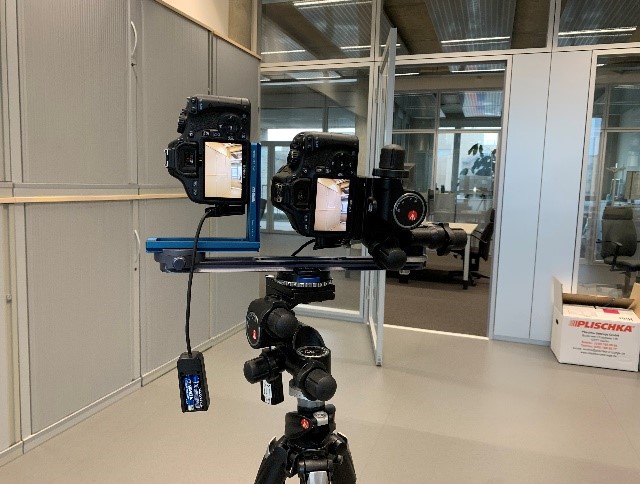}
    \caption{Stereo camera setup with two calibrated Canon EOS 550D.}
    \label{fig:camera-setup}
\end{figure}

\begin{figure}[ht]
    \centering
    \includegraphics[width=0.5\linewidth]{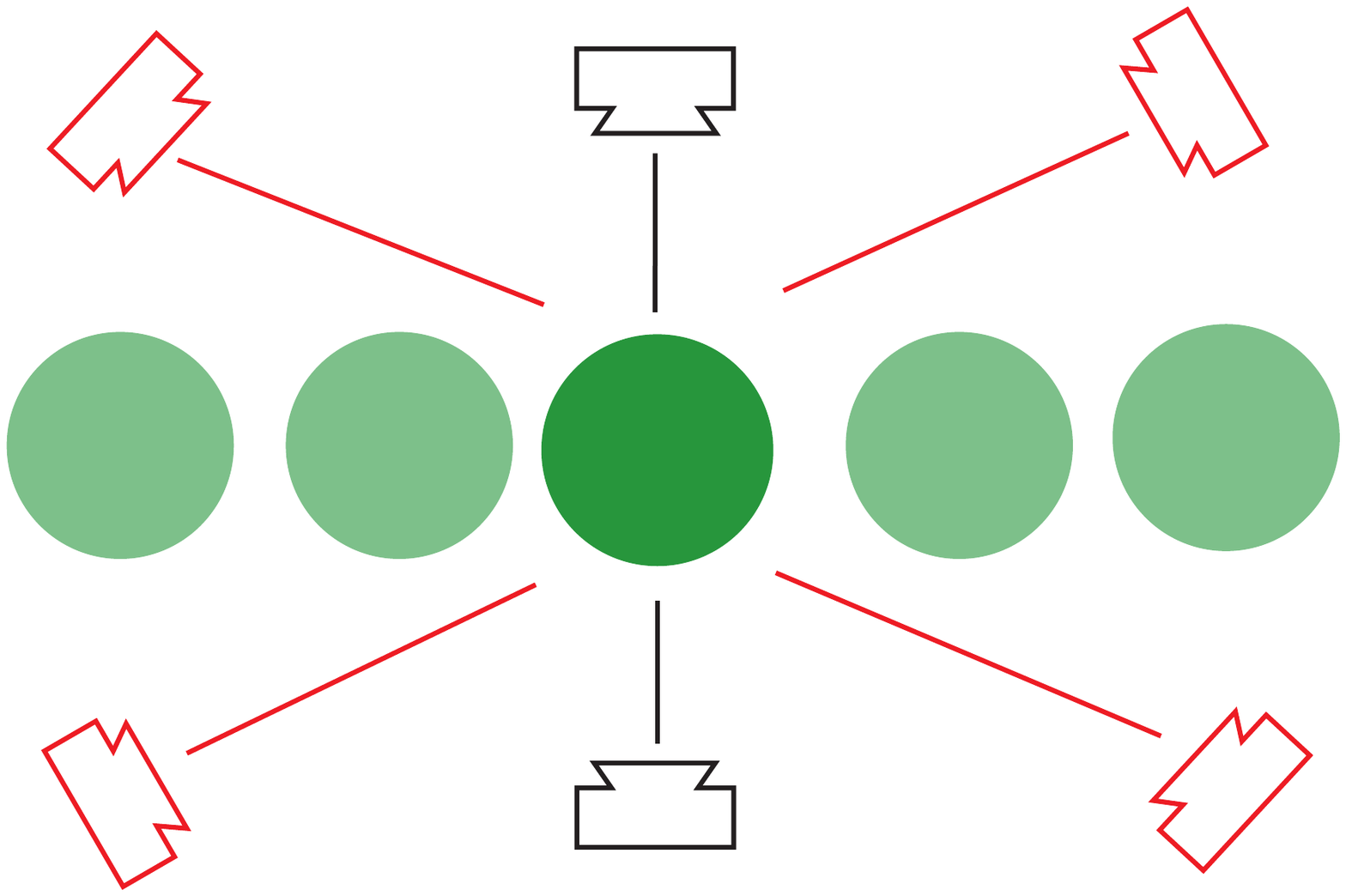}
    \caption{Positions per tree in a row on an orchard using a stereo setup per position}
    \label{fig:aquisition_bbobst}
\end{figure}

At orchards of Berlin-Brandenburg Obst (BB Obst), we collected images of the same 33 Jonagold trees on 18 occasions throughout 2022. BB Obst cultivates fruits, is a fruit retailer and is located in Wesendahl near Berlin. Next to cherries, pears, plums, and strawberries many apple varieties are grown in spindle form. We constructed a stereo imaging setup using two Canon EOS 550D cameras with 20mm lenses, which were triggered simultaneously (see figure \ref{fig:camera-setup}). This setup is calibrated using the calibration object and method described in \cite{eisert_model-based_2002}. Each tree was captured from both sides of the row including front, left, right, and steep angled views, as well as a circular movement around the first tree of the row (Figure \ref{fig:aquisition_bbobst}). Some images were taken in windy conditions, which introduced tree deformations.

Another part of the dataset was collected at the "Sächsisches Landesamt für Umwelt, Landwirtschaft und Geologie" (LfULG) orchard, located in Pillnitz near Dresden. The orchard is used for fruit cultivation research and educational purposes. Many different apple varieties are grown in rows of about 70\unit{m} length, with trees spaced 1\unit{m} apart in spindle form and 3\unit{m} between rows.
For a subset of trees, six images per tree were taken from different angles, as described in the previous section using a smartphone camera. Each tree is referenced by row number and tree position within the row. Data collection took place at various dates in 2023, covering stages from blossom to ripe apples. In 2024, the same trees were documented once more, specifically with the apple count. Additional data was collected from other rows, capturing two images per tree using a smartphone. These images were taken by walking along the rows in one direction and back on the other side, ensuring that the first and last images in the subset correspond to the front and back views of the first tree. In total, the Dresden subset consist of 669 labeled images. In 2024, the orchard experienced a significant loss in fruit due to late frost.

\begin{figure}[ht]
    \centering
    \includegraphics[keepaspectratio, width=0.8\columnwidth]{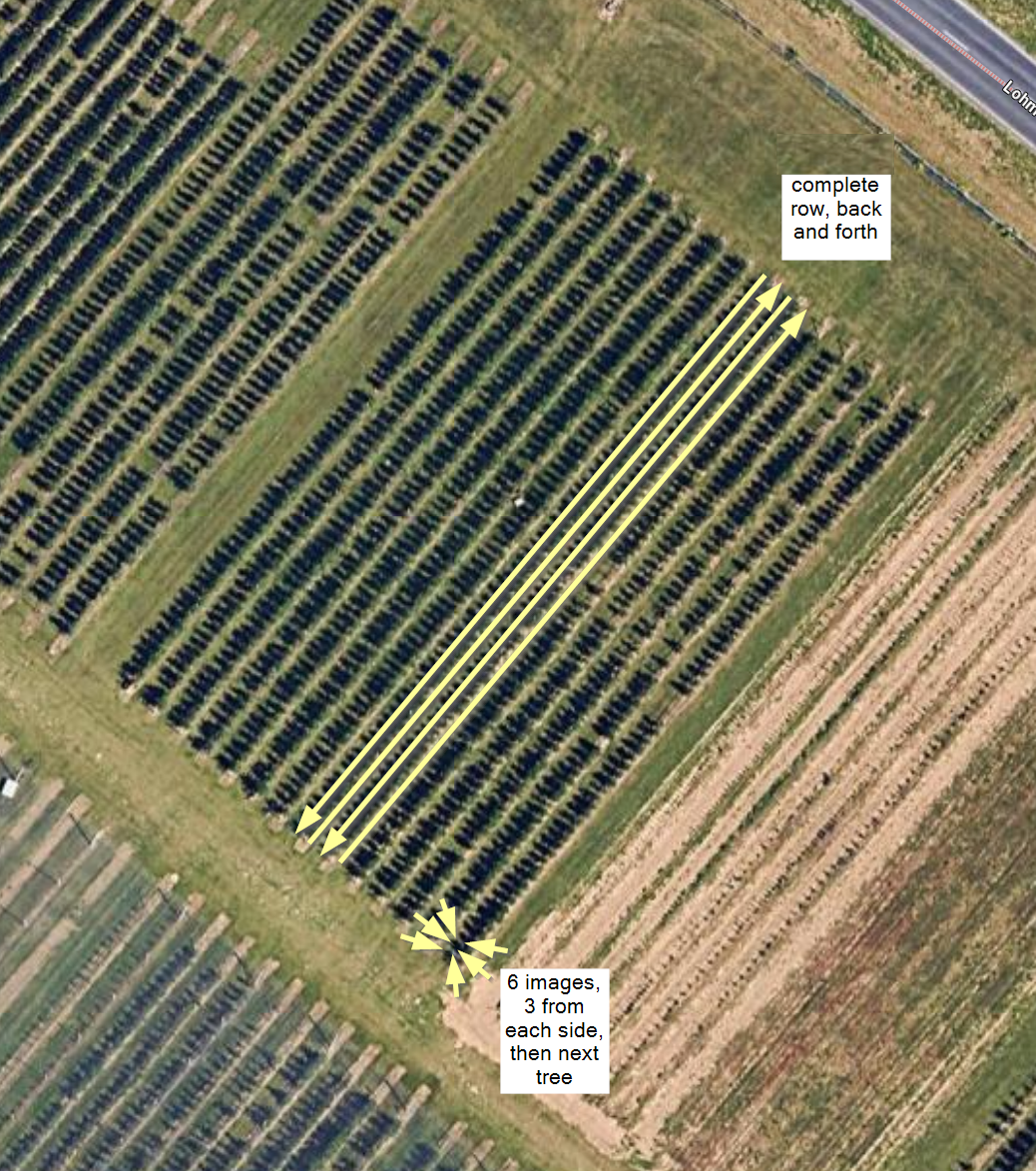}
    \caption{Data acquisition in Pillnitz}
    \label{fig:dataaquisitionpillnitz}
\end{figure}

For 10 trees which are included in the images taken in Pillnitz in 2024, a manual reference count for the number of apples on the trees was performed. It was done without harvest, all apples remained on the trees. Information about the counted number of apples is provided with the dataset.

\subsection{Data Annotation}

The dataset was annotated for apple recognition using an in-house developed annotation software, designed to streamline the labeling process through a combination of AI-assisted automation and human verification. To facilitate annotation, we trained YOLOv8 on a manually labeled subset of 108 images in non-stereo and 70 stereo images, combined with the full MinneApple dataset. This initial model generated automatic annotations for the remaining, unlabeled images. The annotations consist of bounding boxes drawn around individual apples and were created only for images representing BBCH growth stages five to nine, as these stages are most relevant for fruit visibility and detection.

To improve the quality of the annotations, human annotators reviewed and corrected the AI-generated labels for 70 images of the stereo-format, and for 777 of the non-stereo images of the dataset. The final annotations consist of bounding boxes around apples in the images, stored in Darknet/YOLO format used by the Ultralytics model training framework. This semi-automated annotation approach significantly reduced the time and effort required for labeling while maintaining a high level of accuracy.

\subsection{Growth Stage Classification}

From each date in the dataset, a subset of images was randomly selected for growth stage categorization based on the extended BBCH scale for pome fruit. This classification was performed manually by an expert from LfULG Sachsen. Local variations within different sections of the orchard and variability in individual tree development result in different secondary growth stages of trees outside the random sample. Table \ref{tab:BBCH_statistics_in_dataset} shows an overview of all stages contained in the AppleGrowthVision dataset, sorted by scale. This information allows for the selection of subsets of the dataset, if a specific growth stage is of special interest.

\begin{table}
  \centering
  \begin{tabular}{@{}lll@{}}
    \toprule
    BBCH & date & location \\
    \midrule
    01   & 22.03.2023 & BB Obst \\
    \multirow{2}{*}{03}  & 11.03.2022 & BB Obst \\ 
                        & 25.03.2022 & BB Obst \\
    \midrule                     
    54 & 01.04.2022 & BB Obst \\
    55 & 08.04.2022 & BB Obst \\
    56  & 21.04.2023 & BB Obst \\
    57  & 29.04.2022 & BB Obst \\
    \midrule
    60  & 04.05.2023 & BB Obst \\
    61/65 & 25.04.2023 & Pillnitz \\
    \multirow{2}{*}{65}  & 04.05.2023 & Pillnitz \\
                         & 06.05.2022 & BB Obst \\
    \multirow{2}{*}{67}   & 13.05.2022 & BB Obst \\
                          & 16.05.2023 & BB Obst \\
    \midrule
    71/72  & 24.05.2023 & Pillnitz \\
    72  & 30.05.2022 & BB Obst \\
    73  & 13.06.2022 & BB Obst \\
    74  & 27.06.2022 & BB Obst \\
    77  & 03.08.2022 & BB Obst \\
    \midrule
    81  & 23.08.2023 & Pillnitz \\
    85  & 26.08.2024 & Pillnitz \\
    \multirow{2}{*}{87} & 06.09.2022 & BB Obst \\  
                        & 19.09.2023 & Pillnitz \\
    \midrule
    91  & 02.11.2022 & BB Obst \\
    \bottomrule
  \end{tabular}
  \caption{extended BBCH stages included in dataset}
\label{tab:BBCH_statistics_in_dataset}
\end{table}

\begin{figure}[ht]
    \centering
    \begin{subfigure}{0.3\linewidth}
        \centering
        \includegraphics[width=\linewidth,trim={6.3cm 0 6cm 0},clip]{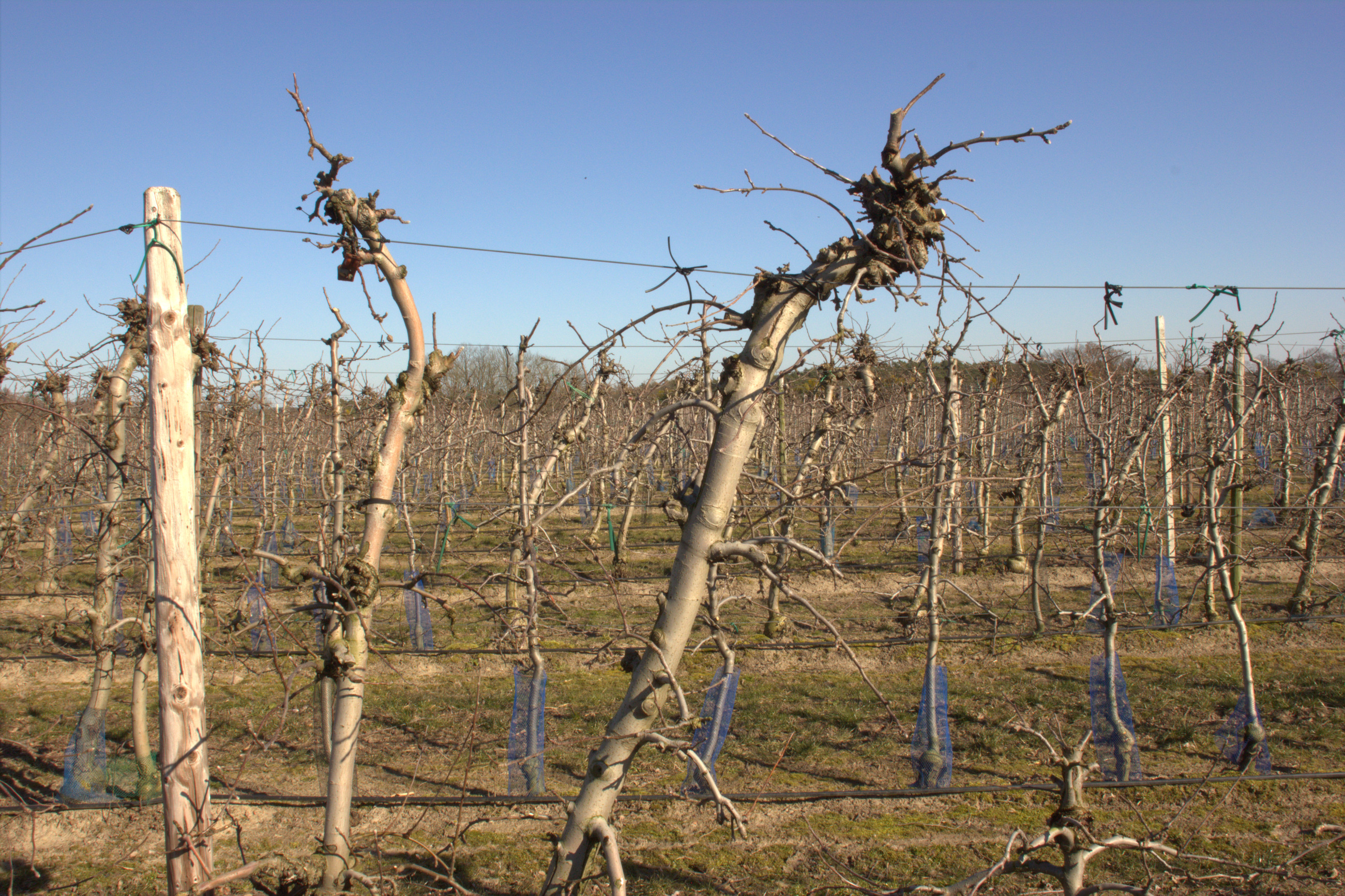}
        \caption{2022-03-11, bud development}
    \end{subfigure}
    \hspace{5pt}
    \begin{subfigure}{0.3\linewidth}
        \centering
        \includegraphics[width=\linewidth]{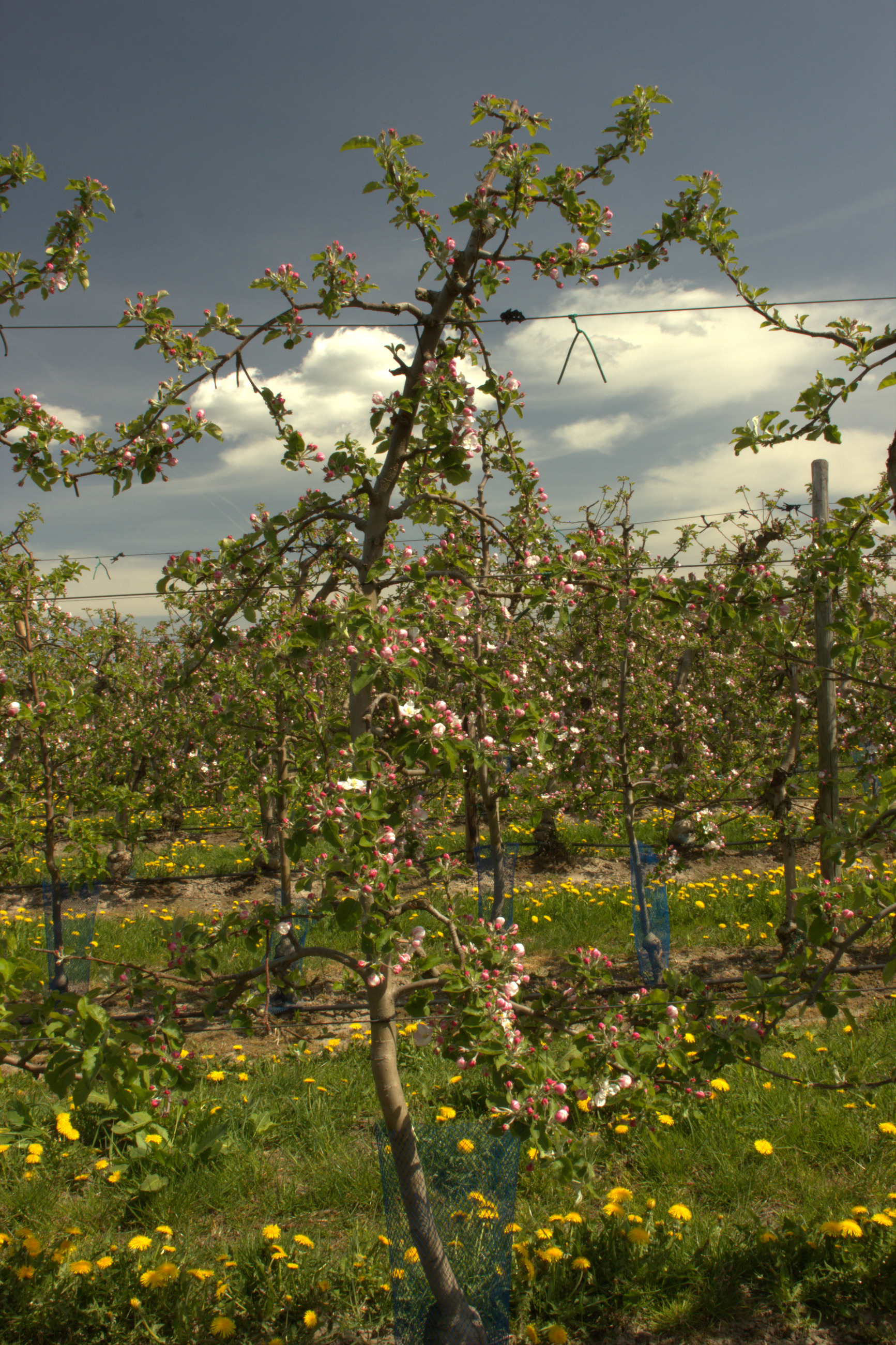}
        \caption{2022-04-20, inflorescence emergence}
    \end{subfigure}
    \hspace{5pt}
    \begin{subfigure}{0.3\linewidth}
        \centering
        \includegraphics[width=\linewidth]{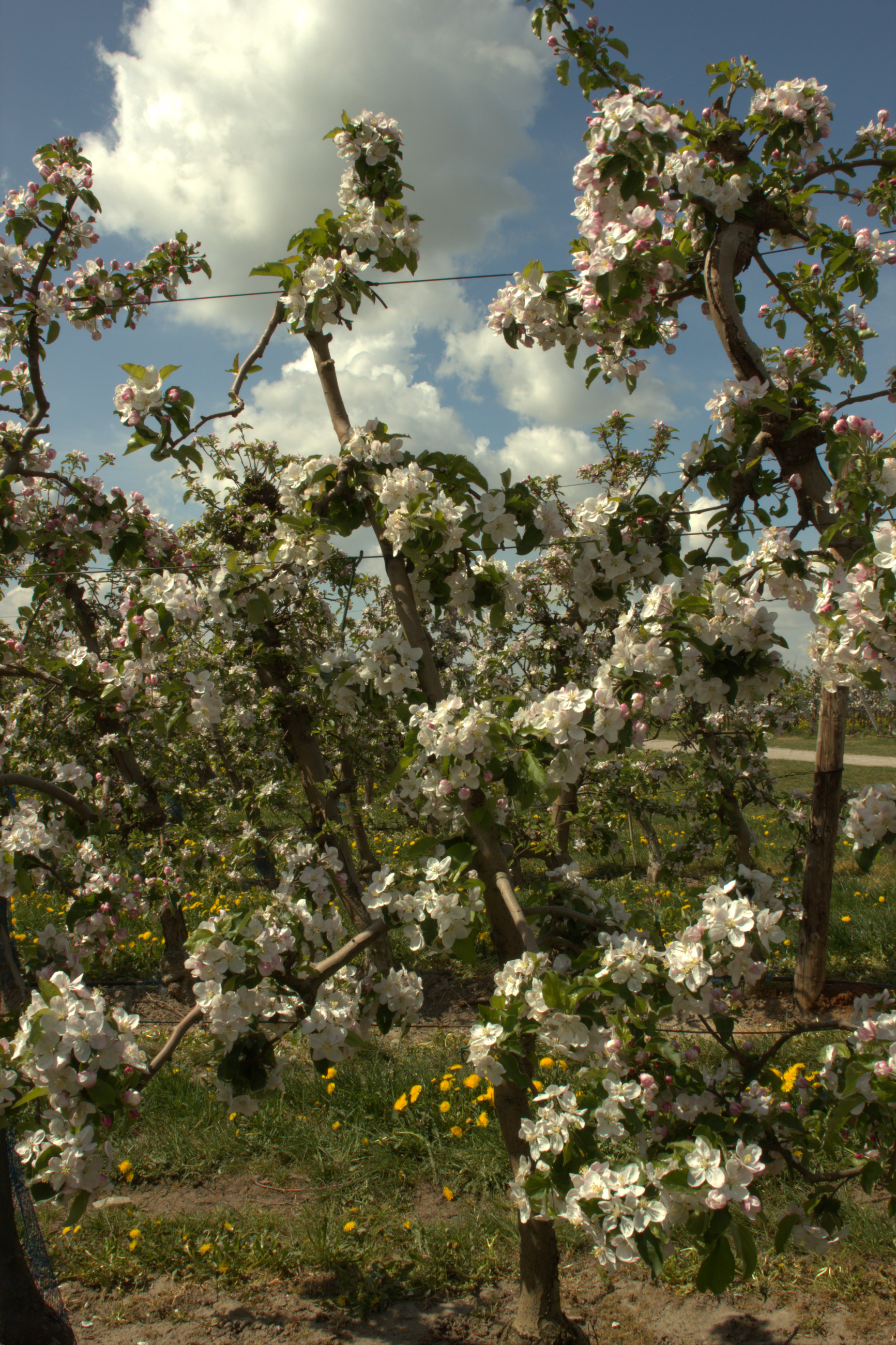}
        \caption{2202-05-06, flowering}
    \end{subfigure}
    
    \vspace{10pt} 

    \begin{subfigure}{0.3\linewidth}
        \centering
        \includegraphics[width=\linewidth]{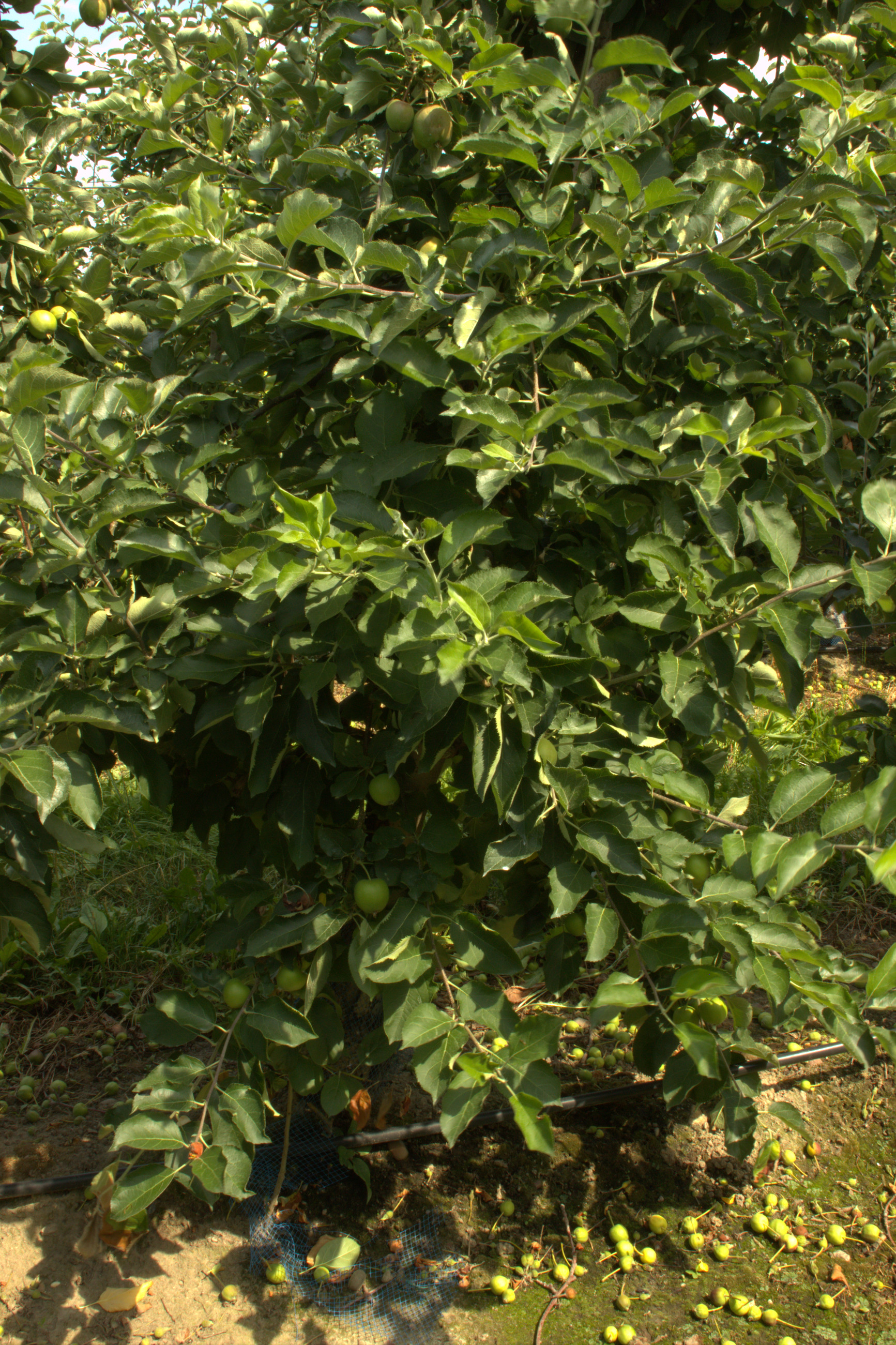}
        \caption{2022-06-27, development of fruit}
    \end{subfigure}
    \hspace{5pt}
    \begin{subfigure}{0.3\linewidth}
        \centering
        \includegraphics[width=\linewidth]{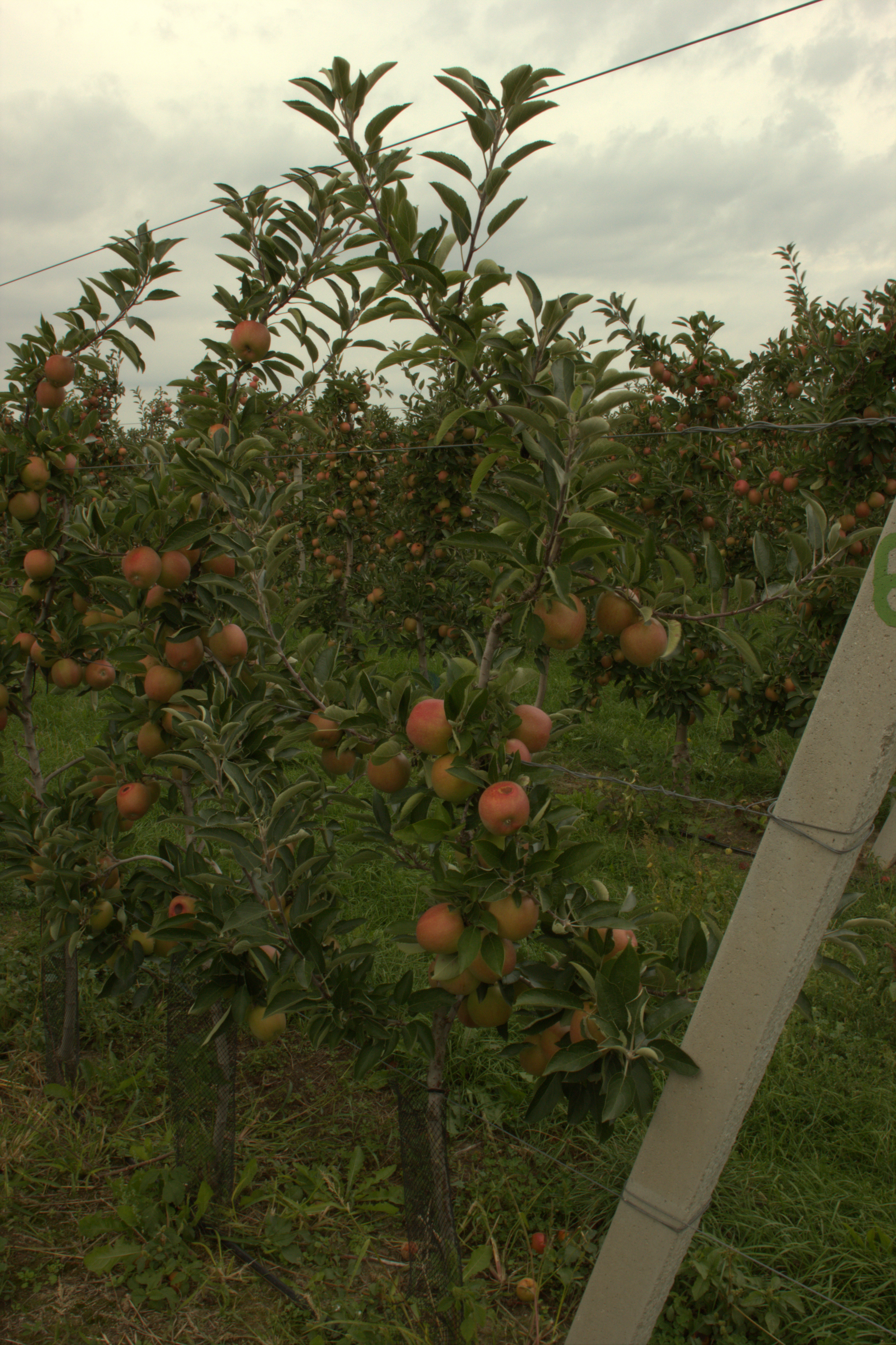}
        \caption{2022-09-06, maturity of fruit and seed}
    \end{subfigure}
    \hspace{5pt}
    \begin{subfigure}{0.3\linewidth}
        \centering
        \includegraphics[width=\linewidth]{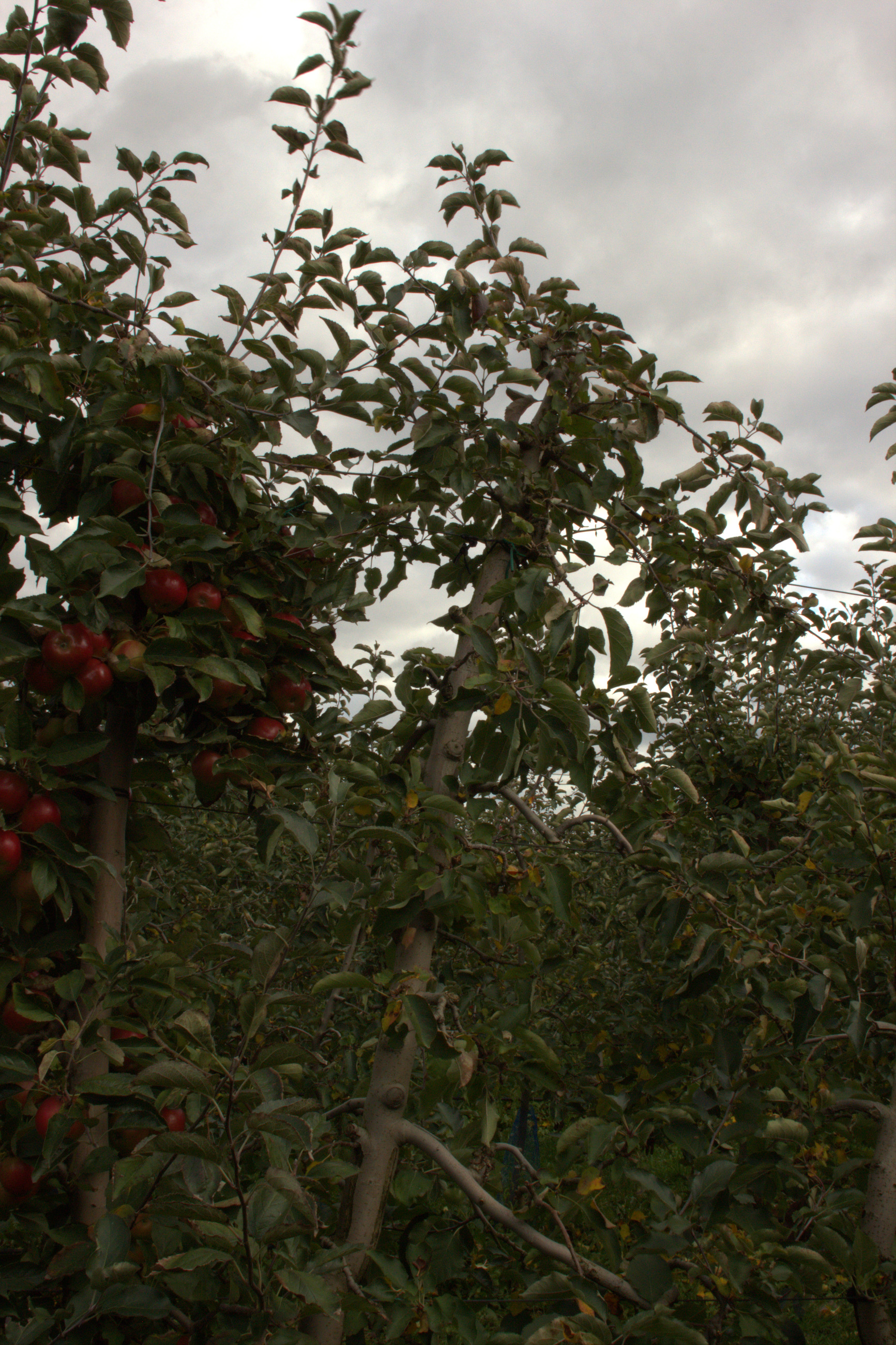}
        \caption{2022-11-02, beginning of dormacy}
    \end{subfigure}

    \caption{Example images for each existing growth stage in BBCH scale notation from the dataset from growth cycle 2022 of the Brandenburg subset.}
    \label{fig:bbch_images}
\end{figure}

Figure \ref{fig:bbch_images} shows images from each of the six stages of the Brandenburg subset of the AppleGrowthVision dataset.

\section{Dataset Evaluation}
\label{sec:statistics}

We conducted a series of experiments aimed at assessing the dataset’s effectiveness in two key areas. Training and validating object detection models for fruit localization and growth stage prediction, and enabling 3D reconstruction of orchard scenes from calibrated stereo imagery. These evaluations provide insight into the dataset’s quality, robustness, and relevance for advancing both phenological monitoring and spatial analysis in orchard environments.

\subsection{Apple Detection: YOLOv8, Faster R-CNN}

We evaluated the impact of our dataset on improving fruit detection models in orchard environments. We used YOLOv8 and Faster R-CNN \cite{ren2015faster}, two widely adopted object detection architectures, to assess how different training configurations affect performance. For further evaluation, applying a transformer-based approach would be valuable, as transformer architectures have shown promising results when using large amounts of image data \cite{dosovitskiy2021, carion2020} and could enhance detection accuracy in dense, occluded orchard scenes.

Our evaluation focused on how extending publicly available datasets like MinneApple and MAD improves fruit detection accuracy. The improvement in apple detection accuracy is assessed on the MAD and MinneApple validation sets. The models were trained on various dataset combinations and evaluated using Average Precision (AP), Average Recall (AR), mean Average Precision at IoU thresholds of 0.5 (mAP@0.5) and 0.5:0.95 (mAP@0.5:0.95), and F1 score.

Table \ref{tab:experiment_results} summarizes the results of our experiments. We observed that augmenting the MinneApple dataset with AppleGrowthVision training data, which contains manually optimized annotations, improves detection performance of YOLOv8 by 5.87 \% in terms of F1-score. Furthermore, incorporating the MAD dataset enhances model performance by an additional 1.82 \%, demonstrating the benefits of dataset diversity and additional training data. Training solely on MinneApple resulted in the lowest performance across four of the five metrics when compared to configurations that included all three datasets, indicating limited generalization to more diverse data. The best results relative to training on MinneApple alone were achieved when all three datasets were combined. This configuration yielded the highest AR- and mAP-values, and F1-score, highlighting the value of combining multiple complementary datasets to build robust object detection models for real-world orchard environments.

The experiments with Faster R-CNN followed a similar structure but revealed important differences in behavior. When trained on MinneApple and MAD, Faster R-CNN achieved an AP-, AR-, mAP- and F1-scores that were lower than those of YOLOv8 under the same conditions. Adding AppleGrowthVision to the training set improved AP-, AR- and F1-score, demonstrating that Faster R-CNN also benefits from dataset diversity. Interestingly, the best performance across three of the five metrics was achieved when trained solely on MinneApple and AppleGrowthVision, while the worst results occurred when training only on MinneApple and MAD. This indicates higher sensitivity of Faster R-CNN to scene variability and potential overfitting when trained on a limited combination of datasets. Notably, performance recovered when all three datasets were combined in experiment (2): AP increased by 31.74 \%, AR by 30.38 \%, mAP@0.5 by 42.12 \%, mAP@.5:.95 by 57.21 \%, and F1 by 31.06 \% compared to experiment (3). Although this was still lower than YOLOv8 in the same setup, it confirms that dataset richness contributes positively to generalization, even for models that are more sensitive to training-test domain shifts.

These findings suggest that adding diverse and phenologically rich datasets significantly improves fruit detection performance. While YOLOv8 consistently outperforms Faster R-CNN in overall accuracy and robustness, both architectures benefit from integrating multiple datasets. This underscores the importance of leveraging complementary data sources to build resilient object detection models for complex orchard environments.

\begin{table}
  \centering
  \setlength{\tabcolsep}{3pt} 
  \begin{tabular}{@{}p{2cm} p{2cm} c c c c@{}}
    \toprule
     Model & Metric & (1) & (2) & (3) & (4) \\
    \midrule
    YOLOv8 & AP & .814 & \textbf{.822} & .81 & .811 \\
     & AR & .637 & .701 & .716 & \textbf{.733} \\
     & mAP@.5 & .753 & .793 & .805 & \textbf{.81} \\  
     & mAP@.5:.95 & .424 & .453 & .456 & \textbf{.464} \\
     & F1 & .715 & .757 & .76 & \textbf{.77} \\
    \midrule
    Faster & AP & .766 & \textbf{.789} & .564 & .743 \\
    R-CNN & AR & .613 & .639 & .497 & \textbf{.648} \\
     & mAP@.5 & .694 & \textbf{.721} & .47 & .668 \\  
     & mAP@.5:.95 & .352 & .345 & .194 & .305 \\
     & F1 & .681 & \textbf{.706} & .528 & \textbf{.692} \\
    \bottomrule
  \end{tabular}
  \caption{Comparison of model performance using different dataset combinations: (1) MinneApple, (2) MinneApple + AppleGrowthVision, (3) MinneApple + MAD, (4) MinneApple + AppleGrowthVision + MAD.}
  \label{tab:experiment_results}
\end{table}

\subsection{BBCH-scale classification}
Following the experiments from \cite{nguyen_bbch_2025}, we finetuned a VGG16 \cite{simonyan_vgg_2015}, a ResNet152 \cite{resnet}, a DenseNet201 \cite{densenet}, 
and a MobileNetv2 \cite{mobilenet}, which were all pretrained on ImageNet1k \cite{you_imagenet_2018}. We measured the performance in accuracy, precision, recall, and F1 score. Our data was split randomly using 80\% training, 10\% validation, and 10\% test data. Table \ref{tab:bbch_classification_results} shows that each method is able to easily distinguish between each principal BBCH stage. A direct comparison to \cite{nguyen_bbch_2025} is not applicable since they focused on secondary BBCH growth stages. Nevertheless, several agricultural tasks rely on the fine-grained resolution provided by the secondary BBCH growth stages. For example, fungicide application schedules are strongly tied to phenological stages such as petal fall, which is the last stage of flower development \cite{ellis1998}. Similarly, accurate yield estimation depends on distinguishing how far fruit development has progressed within BBCH stage 7, as early fruit set (e.g., BBCH 71) and near-mature fruit (e.g., BBCH 77) indicate very different outcomes. Finally, determining optimal harvest time often depends on secondary stages like BBCH 85, which marks full ripeness, rather than simply recognizing that the fruit is in the ripening phase \cite{Meier1994}.

\begin{table}
  \centering
  \setlength{\tabcolsep}{3pt} 
  \begin{tabular}{@{}p{2.5cm} c c c c@{}}
    \toprule
     Model & Accuracy & Precision & Recall & F1 Score \\
    \midrule
    VGG16 & 0.988 & 0.988 & 0.988 & 0.988 \\
    ResNet152 & 0.997 & 0.997 & 0.997 & 0.997  \\
    DenseNet201 & 0.999 & 0.999 & 0.999 & 0.999  \\
    MobileNetv2 & 0.999 & 0.999  & 0.999 & 0.999 \\
    \bottomrule
  \end{tabular}
  \caption{Comparison of model performance on test data for BBCH stage classification.}
  \label{tab:bbch_classification_results}
\end{table}

\subsection{Multi-View Stereo Reconstructions}

\begin{figure}[t]
    \centering
    \includegraphics[keepaspectratio, width=\linewidth]{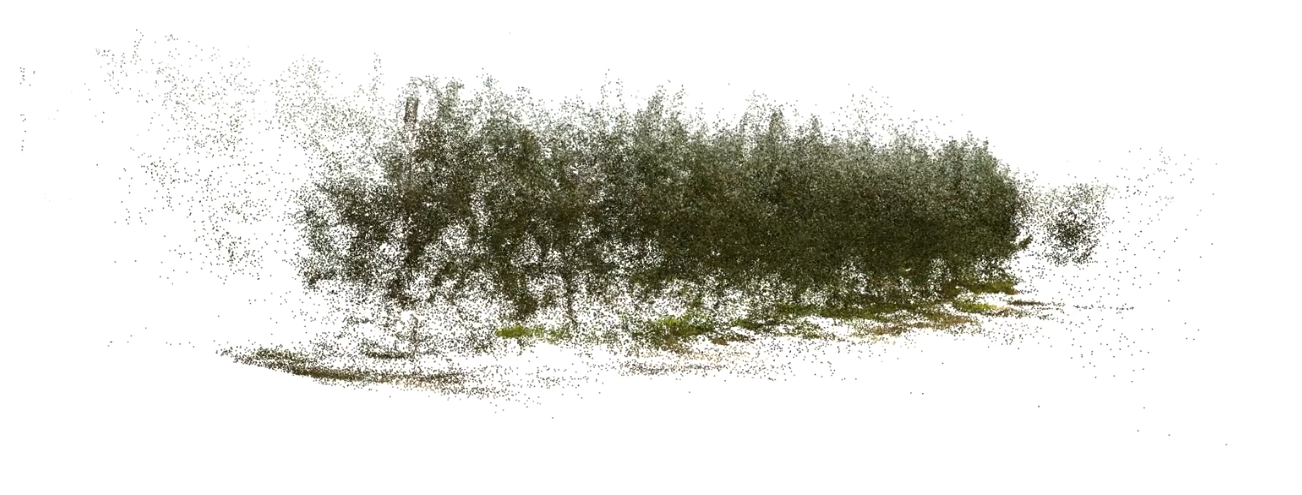}
    \caption{Example of a reconstructed orchard scene of apple trees from BB Obst.}
    \label{fig:example-reconstruction}
\end{figure}

Calibrated stereo vision data is used to extract 3D reconstructions of scenes. Reconstructions enable automated phenological measures and growth monitoring. These applications depend on the number of captured images per scene and the absence of motion for the target objects. Wind that bends trees is one big challenge in outside scenes. For CherryPicker 250 images were used in almost windless conditions to retrieve point clouds for one specific tree \cite{meyer_cherrypicker_2023}. For common representations, such as point clouds \cite{meyer_cherrypicker_2023}, NeRFs \cite{fruitnerf2024}, and Gaussian Splattings \cite{kerbl_3d_2023}, it is assumed that camera positions are given. With our calibrated data, state-of-the-art methods like COLMAP \cite{schonberger_structure--motion_2016} and  Mast3r \cite{leroy2024grounding} are not able to locate the cameras in these large and complex scenes. SIFT \cite{lowe_object_1999}, used by COLMAP, is not capable to describe features meaningfully to distinct the feature points of orchard scenes. Instead we extracted DISK \cite{tyszkiewicz_disk_2020} features, matched them with LightGlue \cite{lindenberger_lightglue_2023} within the HLOC framework \cite{sarlin2019coarse, sarlin2020superglue}, and applied low level adjustments in COLMAP to reconstruct some of these large scenes. In COLMAP we provided an initial pair using the extrinsic calibration priors of one image pair. Then we reconstructed the scene on the initial pair and only the left camera images, adding the right camera images in a second step to optimize computation speed. These reconstructions show that reconstructions of orchards with only stereo images is possible and serve as a baseline for further research on reconstructing these challenging scenes to derive point clouds, NeRFs, and Gaussian Splattings. One reconstructed scene is shown in figure \ref{fig:example-reconstruction}.

\section{Conclusion and Discussion}
\label{sec:discussion}
This work presents AppleGrowthVision, a novel dataset designed to advance computer vision and 3D applications in precision agriculture, particularly for fruit detection and phenological analysis in apple orchards. Existing datasets, such as MinneApple and MAD, have contributed significantly to orchard-based computer vision research but lack agriculturally validated growth stages and comprehensive 3D reconstruction capabilities. Our dataset fills this gap by providing high-resolution stereo images collected over an entire growth cycle with expert-validated BBCH growth stages, enabling more precise modeling of fruit development.

To begin a simple initial evaluation of AppleGrowthVision’s effectiveness, we conducted eight experiments using YOLOv8 and Faster R-CNN, comparing different training configurations to assess improvements in fruit detection performance. The results show that integrating AppleGrowthVision into training enhances detection accuracy across multiple performance metrics. Secondly, we fine-tuned state-of-the-art classification models to predict BBCH growth stages, enabling high-precision support for decision-making and planning in agriculture. All growth stages present in our dataset can be predicted with an accuracy exceeding 95 \%. Thirdly, we used calibrated stereo vision data to reconstruct 3D orchard scenes, supporting automated phenological monitoring despite challenges such as wind and dense vegetation. Since traditional feature matching methods like SIFT failed in these complex environments, we implemented a custom pipeline using DISK features, LightGlue, and COLMAP with calibration priors. This approach successfully produced multi-view stereo reconstructions, providing a baseline for future work on point clouds, NeRFs, and Gaussian Splattings.

While AppleGrowthVision represents an advancement in orchard-based computer vision research, several challenges remain. Automated annotation is not yet sufficient for reliable crop monitoring and yield prediction for the entire data set. Therefore, further improvements in automatic labeling techniques are necessary to enhance annotation accuracy and consistency across the dataset. On the one hand, many agricultural tasks depend on the detailed resolution of secondary BBCH growth stages. For instance, fungicide application is timed to stages like petal fall \cite{ellis1998}, yield estimation varies between early (BBCH 71) and late fruit set (BBCH 77), and optimal harvest decisions rely on stages like BBCH 85, which marks full ripeness \cite{Meier1994}. Therefore, a fully annotated dataset of secondary growth stages is essential for developing reliable and useful models in precision agriculture. On the other hand, further research is needed to determine the level of annotation accuracy—e.g., benchmark F1-scores—required for primary growth stages to develop reliable agricultural technology. This would also help assess whether the annotations of AppleGrowthVision meet those requirements.

To elevate AppleGrowthVision to a widely used benchmark dataset, further validation is needed using both high-capacity and efficient models. Transformer-based architectures such as DETR \cite{carion2020} or Vision Transformers (ViT) \cite{dosovitskiy2021} have demonstrated strong performance in modeling complex spatial relationships, particularly when trained on large-scale datasets. Applying such models to AppleGrowthVision could improve detection accuracy in dense, occluded orchard scenes and provide insights into scalability across large and variable datasets. At the same time, ensuring real-world usability requires evaluating lightweight models that enable real-time inference on edge devices. YOLOv8 shows promising results in this regard, balancing accuracy and efficiency. However, further benchmarking with resource-aware models such as YOLO-NAS, MobileNetV3, or EfficientDet \cite{tan2020, terven2024, howard2019} is necessary to assess performance in constrained environments typical for agricultural robotics and embedded systems. 

Despite the benefits of stereo imagery, handling occlusions, scale variations, and fine-grained structure details remains challenging. Additionally, data collection or synthetic augmentation is needed to ensure complete coverage of all BBCH growth stages, including substages, for a more holistic phenological analysis. Finally, combining RGB, multispectral, and depth information could further improve object detection, classification, and phenotypic trait analysis in orchard environments. With further research and refinement, AppleGrowthVision has the potential to serve as a benchmark dataset for fruit detection, growth modeling, and 3D reconstruction in orchard environments, laying the foundation for future advancements in precision agriculture and autonomous farming technologies.

{
    \small
    \bibliographystyle{ieeenat_fullname}
    \bibliography{main}
}


\end{document}